\documentclass[11pt,letterpaper]{article}
\usepackage{ijcnlp2017}
\usepackage{times}
\usepackage{latexsym}
\usepackage{graphicx}
\usepackage{CJK}
\usepackage{algorithm}
\usepackage{float}
\usepackage[noend]{algcompatible}
\usepackage{latexsym}
\usepackage{multirow}
\usepackage{amsmath}

\definecolor{light-gray}{gray}{0.87}

\ijcnlpfinalcopy

\def\ijcnlppaperid{***}

\title{Improving Neural Machine Translation\\through Phrase-based Forced Decoding}

\author{Jingyi Zhang$^{1,2}$, Masao Utiyama$^1$, Eiichro Sumita$^1$\\ \bf{Graham Neubig}$^{3,2}$, Satoshi Nakamura$^2$\\
$^1$National Institute of Information and Communications Technology,  Japan\\
$^2$Graduate School of Information Science,
Nara Institute of Science and Technology,  Japan\\
$^3$Language Technologies Institute, Carnegie Mellon University, USA \\
  {\tt   jingyizhang/mutiyama/eiichiro.sumita@nict.go.jp }\\
  {\tt  gneubig@cs.cmu.edu, s-nakamura@is.naist.jp}
}

\date{}

\begin{document}

\maketitle

\begin{abstract}
Compared to traditional statistical machine translation (SMT), neural machine translation (NMT) often sacrifices adequacy for the sake of fluency.
We propose a method to combine the advantages of traditional SMT and NMT by exploiting an existing phrase-based SMT model to compute the phrase-based decoding cost for an NMT output and then using  this cost to rerank the $n$-best NMT outputs. 
The main challenge in implementing this approach is that NMT outputs may not be in the search space of the standard phrase-based decoding algorithm,
because the search space of phrase-based SMT is limited by the phrase-based translation rule table.
We propose a soft forced decoding algorithm, which can always successfully find a decoding path for any NMT output. 
We show that using the forced decoding cost to rerank the NMT outputs can successfully improve translation  quality  on four different language pairs.
\end{abstract}

\section{Introduction}

Neural machine translation (NMT), which uses  a single  large neural network to model the entire translation process, has recently been shown to outperform traditional statistical machine translation (SMT) such as phrase-based machine translation (PBMT) on several translation tasks \cite{koehn2003statistical,bahdanau2014neural,sennrich-haddow-birch:2016:WMT}.
Compared to traditional SMT, NMT generally produces more fluent translations, but often sacrifices adequacy, such as translating  source words into completely unrelated target words, over-translation or under-translation \cite{koehn2017six}.

There are a number of methods that combine the two paradigms to address their respective weaknesses.
For example, it is possible to incorporate neural features into traditional SMT models to disambiguate hypotheses \cite{neubig15wat,stahlberg-EtAl:2016:P16-2}.
However, the search space of traditional SMT is usually limited by translation rule tables, reducing the ability of these models to generate hypotheses on the same level of fluency as NMT, even after reranking. 
There are also methods that incorporate knowledge from traditional SMT into NMT, such as lexical translation probabilities \cite{arthur-neubig-nakamura:2016:EMNLP2016,he2016improved}, phrase memory \cite{tang2016neural,zhang-EtAl:2017:Long2}, and  $n$-gram posterior probabilities based on traditional SMT translation lattices \cite{stahlberg-EtAl:2017:EACLshort}.
These improve the adequacy of NMT outputs, but do not impose hard alignment constraints like traditional SMT systems and therefore cannot effectively solve all over-translation or under-translation problems.

In this paper, we propose a method that exploits an existing phrase-based translation model to compute the phrase-based decoding cost for a given NMT translation.%
\footnote{In fact, our method can take in the output of \textit{any} up-stream system, but we experiment exclusively with using it to rerank NMT output.}
That is, we force a phrase-based translation system to take in the source sentence and generate an NMT translation.
Then we use the cost of this phrase-based forced decoding to rerank the NMT outputs.
The phrase-based decoding cost will heavily punish completely unrelated translations, over-translations, and under-translations, as they will not be able to be found in the translation phrase table.

 One challenge in implementing this method is that the NMT output may not be in the search space of the  phrase-based  translation model, which is limited by the phrase-based translation rule table.
 To solve this problem, we propose a soft forced decoding algorithm, which is based on the standard phrase-based decoding algorithm and integrates new types of translation rules (deleting a source word or inserting a target word).
 The proposed forced decoding  algorithm  can always successfully  find  a decoding path  and compute a phrase-based decoding cost for any NMT output. 
 Another  challenge is that we need a diverse NMT $n$-best list for reranking. 
 Because beam search for NMT often lacks  diversity in the beam -- candidates   only have slight differences, with most of the words overlapping -- we use a random sampling method to obtain a more diverse $n$-best list.
  
We test the proposed method on English-to-Chinese, English-to-Japanese, English-to-German and English-to-French translation tasks, obtaining large improvements over a strong NMT baseline that already incorporates discrete lexicon features.
  \begin{figure*}[t]
      \center
      \includegraphics[width=0.99\textwidth]{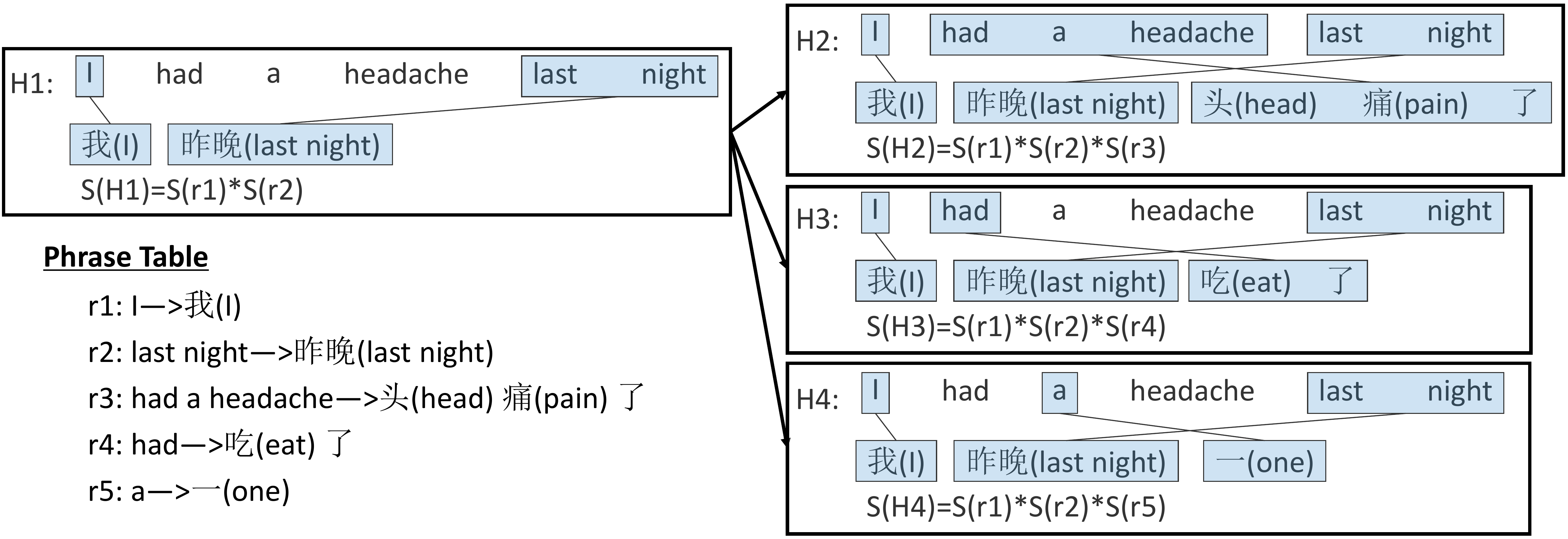}
      \caption{An example of phrase-based decoding.  }
      \label{f1}
    \end{figure*}
\section{Attentional NMT}
Our baseline NMT model is similar to the attentional model of \newcite{bahdanau2014neural}, which includes an encoder, a decoder and an attention (alignment) model.
Given a source sentence $F = \left\{ {{f_1},...,{f_J}} \right\}$, the encoder learns an annotation ${h_j} = \left[ {{{\vec h}_j};{{\mathord{\buildrel{\lower3pt\hbox{$\scriptscriptstyle\leftarrow$}} 
\over h} }_j}} \right]$ for $f_j$ using a bi-directional recurrent neural network. 

The decoder generates the target translation from left to right. The probability of generating next word $e_i$ is,\footnote{$g$, $f$ and $a$ in Equation~\ref{e1}, \ref{e2} and \ref{e4}  are nonlinear, potentially multi-layered, functions.}
\begin{equation}
P_{NMT}\left( {{e_i}|e_1^{i - 1},F} \right) = softmax\left( {g\left( {{e_{i - 1}},{t_i},{s_i}} \right)} \right)
\label{e1}
\end{equation} 
where  $t_i$ is a decoding state for time step $i$, computed by, 
\begin{equation} 
{t_i} = f\left( {{t_{i - 1}},{e_{i - 1}},{s_i}} \right)
\label{e2}
\end{equation}
 $s_i$ is a source representation for time $i$, calculated as,
\begin{equation} 
{s_i} = \sum\limits_{j = 1}^J {{\alpha _{i,j}} \cdot {h_j}} 
\label{e3}
\end{equation} 
where ${\alpha _{i,j}}$ scores  how well the inputs around position $j$ and the output at position
$i$ match, computed as,
\begin{equation} 
{\alpha _{i,j}} = \frac{{\exp \left( {a\left( {{t_{i - 1}},{h_j}} \right)} \right)}}{{\sum\limits_{k = 1}^J {\exp \left( {a\left( {{t_{i - 1}},{h_k}} \right)} \right)} }}
\label{e4}
\end{equation}

As we can see,  NMT only learns an attention (alignment) distribution for each target word over all source words and does not provides exact mutually-exclusive word or phrase level alignments.
As a result, it is known that attentional NMT systems make mistakes in over- or under-translation \cite{cohn-EtAl:2016:N16-1,mi-EtAl:2016:EMNLP2016}.

\section{Phrase-based Forced Decoding for NMT}
\subsection{Phrase-based SMT}

In phrase-based SMT \cite{koehn2003statistical},  a phrase-based translation rule $r$ includes a source phrase, a target phrase and a translation score $S\left( r \right)$. Phrase-based translation rules can be extracted from the word-aligned training set and then used to translate new sentences. Word alignments for the training set can be obtained by IBM models \cite{brown1993mathematics}.

Phrase-based decoding uses a list of translation rules to translate source phrases in the input sentence and generate target phrases from left to right.
A basic concept in phrase-based decoding is hypotheses.
As shown in Figure~\ref{f1}, the hypothesis $H_1$ consists of two rules $r_1$ and $r_2$. The score of a hypothesis $S\left( H \right)$ can be calculated as the product of the scores of all applied rules.\footnote{In actual phrase-based decoding it is common to integrate reordering probabilities in the forced decoding score defined in Equation~\ref{p2}. However, because NMT generally produces more properly ordered sentences than traditional SMT, in this work we do not consider reordering probabilities in our forced decoding algorithm.}
An existing hypothesis can be expanded into a new hypothesis by applying a new rule.
As shown in Figure~\ref{f1}, $H_1$ can be expanded into $H_2$, $H_3$ and $H_4$.
$H_2$ cannot be further expanded, because it covers all source words, while $H_3$ and $H_4$ can (and must) be further expanded. 
The decoder starts with  an initial empty hypothesis $H_0$ and selects the  hypothesis with the highest score from all completed hypotheses.

During decoding, hypotheses  are stored in stacks. For a source sentence with $J$ words, the decoder builds $J$ stacks. The hypotheses that cover $j$ source words are stored in stack $s_j$. The decoder expands  hypotheses in ${s_1},{s_2},...,{s_J}$ in turn as shown in Algorithm~\ref{alg1}.  
 \begin{algorithm}[h]
  \label{a1}
    \caption{Standard phrase-based decoding.}
    \begin{algorithmic}
      \REQUIRE Source sentence $F$ with length $J$
      \ENSURE Translation $E$ and decoding path $D$
        \STATE initialize $H_0$ and ${s_1},{s_2},...,{s_J}$
      \STATE \textsc{Expand}$\left( {{H_{0}}} \right)$
      \FOR {$j=1$ to $J-1$}
    
  \FOR {each hypothesis $H_{jk}$ in $s_j$}

      \STATE \textsc{Expand}$\left( {{H_{jk}}} \right)$

    \ENDFOR

    \ENDFOR
    \STATE select best hypothesis in $s_J$
    \end{algorithmic}
    \label{alg1}
  \end{algorithm}  
 Here, \textsc{Expand}$\left( {{H}} \right)$ is expanding $H$ to get new hypotheses and putting the new hypotheses into corresponding stacks.
For each
stack, a beam of the best $n$ hypotheses is kept to speed up the decoding process.
\subsection{Forced Decoding for NMT}

As stated in the introduction, our goal is not to generate new hypotheses with phrase-based SMT, but instead use the phrase-based model to calculate scores for NMT output.
In order to do so, we can perform \textit{forced decoding}, which is very similar to the algorithm in the previous section but discards all partial hypotheses that  do not match the NMT output.
However,  the NMT output is not limited by the phrase-based rule table, so there may be no decoding path that completely matches the NMT output when using only the phrase-based rules. 

To remedy this problem, inspired by previous work in forced decoding for training phrase-based SMT systems \cite{wuebker-mauser-ney:2010:ACL,wuebker-hwang-quirk:2012:WMT} we propose a soft forced decoding algorithm that can always successfully find a   decoding path for a source sentence $F$ and an NMT translation $E$.

First, we introduce two new types of rules R$_1$ and R$_2$.

\paragraph{R$_1$}A source word $f$ can be translated into a special word $\texttt{null}$. This corresponds to deleting $f$ during  translation. The
  score of deleting $f$  is  calculated as,
  \begin{equation}  
s \left( {f \to \texttt{null}} \right) = \frac{{\text{unalign}\left( f \right)}}{{\left| \mathcal{T} \right|}}
\label{r1}
  \end{equation}
where ${\text{unalign}\left( f \right)}$ is how many times $f$ is unaligned in the word-aligned  training set $\mathcal{T}$ and $\left| \mathcal{T} \right|$ is the number of sentence pairs in $\mathcal{T}$.

\paragraph{R$_2$}A target word $e$ can be translated from a special word $\texttt{null}$, which corresponds to inserting $e$    during translation. The  score of inserting $e$  is  calculated as,
\begin{equation}  
s \left( {\texttt{null} \to e} \right) = \frac{{\text{unalign}\left( e \right)}}{{\left| \mathcal{T} \right|}}
\label{r2}
\end{equation}
where ${\text{unalign}\left( e \right)}$ is how many times $e$ is unaligned in $\mathcal{T}$.

One motivation for Equations~\ref{r1} and~\ref{r2} is that function words usually have high frequencies, but do not have as clear a correspondence with a word in the other language as content words.
As a result, in the training set function words are more often unaligned than content words.
As an example, Table~\ref{times} and Table~\ref{timesfr} show how many times  different words occur and how many times they are unaligned  in the word-aligned training set of English-to-Chinese and English-to-French tasks in our experiments. As we can see, generally there are less unaligned words in the English-to-French task, however, function words are more likely to be unaligned in both tasks.
Based on Equation~\ref{r1} and Equation~\ref{r2}, the scores of   deleting or inserting  ``of" and ``a" will be higher. 
 
 \begin{table}[h] 
               \center
                                                                                                                          
                \begin{tabular}{l|rrrr} 
                \hline 
             
             Words&of&a&practice&water\\
             \hline 
             Occur&1.3M&1.0M&2.2K&29K\\
             Unaligned&0.51M&0.41M&0.25K&3.5K\\
          
                                      \hline
                                         \end{tabular}
                                                                                                                                                                                                                                                             
                   \caption{The number of times that words occur in the English-to-Chinese training corpus and the number of times that they are unaligned. }
                                    \label{times}
                  \end{table} 

\begin{table}[h] 
               \center
                                                                                                                          
                \begin{tabular}{l|rrrr} 
                \hline 
             
             Words&of&a&practice&water\\
             \hline 
             Occur&1.7M&0.83M&8.8K&7.4K\\
             Unaligned&0.16M&0.12M&0.38K&0.19K\\
          
                                      \hline
                                         \end{tabular}
                                                                                                                                                                                                                                                             
                   \caption{The number of times that words occur in the English-to-French training corpus and the number of times that they are unaligned. }
                                    \label{timesfr}
                  \end{table} 
In our forced decoding, we choose to model the score of each translation rule that exists in the phrase table as the product of direct and inverse phrase translation probabilities.
To make sure that the scale of the scores for R$_1$ and R$_2$ match the other phrase (which are the product of two probabilities), we use the square of the score in Equation~\ref{r1}/\ref{r2}  as the rule score for R$_1$/R$_2$.

         \begin{algorithm}[t]
          \label{a1}
            \caption{Forced phrase-based decoding.}
            \begin{algorithmic}
              \REQUIRE Source sentence $F$ with length $J$ and translation $E$ with length $I$
              \ENSURE Decoding path $D$
              \STATE initialize $H_0$ and ${s{'_1}},{s{'_2}},...,{s{'_I}}$
              \STATE \textsc{Expand}$\left( {{H_{0}}} \right)$
                  \STATE expand $H_{0}$ with rule \texttt{null}$\to e_{1}$ 
              \FOR {$i=1$ to $I-1$}          
          \FOR {each hypothesis $H_{ik}$ in $s{'_i}$}
              \STATE \textsc{Expand}$\left( {{H_{ik}}} \right)$
              \STATE expand $H_{ik}$ with rule \texttt{null}$\to e_{i+1}$ 
            \ENDFOR
            \ENDFOR
               \FOR {each hypothesis $H_{Ik}$ in $s{'_I}$}
                          \STATE update  $S\left( {{H_{Ik}}} \right)$ for uncovered source words
                        \ENDFOR
                           \STATE select best hypothesis in $s{'_I}$
            \end{algorithmic}
            \label{alg2}
          \end{algorithm}

Algorithm~\ref{alg2} shows the forced decoding algorithm that integrates the new rules.
  Because the translation $E$ is given for the forced decoding algorithm, the proposed forced decoding algorithm keeps $I$ stacks, where $I$ is the length of $E$. In other words, the stack size is corresponding to the target word size during forced decoding while the stack size is corresponding to the source word size  during standard phrase-based decoding. The stack $s{'_i}$ in Algorithm~\ref{alg2} 
contains all hypotheses in which the first $i$ target words have
been generated. 
We expand  hypotheses in ${s{'_1}},{s{'_2}},...,{s{'_I}}$ in turn. When expanding a hypothesis $H_{old}$ in $s{'_i}$, besides expanding it  using the original rule table \textsc{Expand}$\left( {{H_{old}}} \right)$,\footnote{The new introduced word inserting/deleting rules are not used when
performing \textsc{Expand}$\left( {{H_{old}}} \right)$.} we also expand $H_{old}$ by inserting the next target word $e_{i+1}$ at the end of $H_{old}$ to get an additional hypothesis $H_{new}$ and put $H_{new}$ into $s{'_{i+1}}$. 
For a final
  hypothesis in  stack $s{'_I}$, it may not cover all source words. We update its  score by  translating
    uncovered words into $\texttt{null}$.

Because different decoding paths can generate the same final translation, there can be different decoding paths that fit the NMT translation $E$.
We use the score of the single decoding path with the highest decoding score  as the forced decoding  score for $E$.

\section{Reranking NMT Outputs with  Phrase-based Decoding  Score}

We rerank the $n$-best NMT outputs using the phrase-based forced decoding  score according to Equation~\ref{rerank}. 
\begin{equation}\small 
\log P\left( {E|F} \right) = {w_1} \cdot \log {P_n}\left( {E|F} \right) + {w_2} \cdot \log {S_d}\left( {E|F} \right)
\label{rerank}
\end{equation}
where ${P_n}\left( {E|F} \right)$ is the original NMT translation probability as calculated by Equation~\ref{e1};
\begin{equation}\small 
{P_n}\left( {E|F} \right) = \prod\limits_{i = 1}^I {P_{NMT}\left( {{e_i}|e_1^{i - 1},F} \right)} 
\label{p1}
\end{equation}
 ${S_d}\left( {E|F} \right)$ is  the forced decoding  score, which is the  score of the  decoding path $\hat D$ with the highest decoding  score as described above; 
 \begin{equation}\small 
 {S_d}\left( {E|F} \right) = \prod\nolimits_{r \in \hat D} {S\left( r \right)} 
 \label{p2}
 \end{equation}
 $w_1$ and $w_2$ are weights that can be tuned on the $n$-best list of the development set.

The easiest way to get an $n$-best list for NMT is by using the $n$-best translations from beam search, which is the standard decoding algorithm for NMT.
While beam search is likely to find the highest-scoring hypothesis, it often lacks diversity in the beam: candidates   only have slight differences, with most of the words overlapping.
In order to obtain a more diverse list of hypotheses for reranking, we additionally augment the 1-best hypothesis discovered by beam search with translations sampled from the NMT conditional probability distribution.

The standard method for sampling hypotheses in NMT is ancestral sampling, where we randomly select a word from the vocabulary according to $P_{NMT}\left( {e_i|e_1^{i - 1},F} \right)$ \cite{shen-EtAl:2016:P16-1}.
This will make a diverse list of hypotheses, but may reduce the probability of selecting a highly scoring hypothesis, and the whole $n$-best list may not contain any candidate with better translation quality than the standard beam search output.

Instead, we take an alternative approach that proved empirically better in our experiments: at each time step $i$, we use sampling to randomly select the next word from $e'$ and $e''$ according to Equation~\ref{random}. Here, $e'$ and $e''$ are the two target words with the highest probability according to Equation~\ref{e1}. 
\begin{equation}\large
\begin{array}{l}
{P_{rdm}}\left( {e'} \right) = \frac{{P_{NMT}\left( {e'|e_1^{i - 1},F} \right)}}{{P_{NMT}\left( {e'|e_1^{i - 1},F} \right) + P_{NMT}\left( {e''|e_1^{i - 1},F} \right)}}\\
{P_{rdm}}\left( {e''} \right) = \frac{{P_{NMT}\left( {e''|e_1^{i - 1},F} \right)}}{{P_{NMT}\left( {e'|e_1^{i - 1},F} \right) + P_{NMT}\left( {e''|e_1^{i - 1},F} \right)}}
\end{array}
\label{random}
\end{equation}
The sampling process ends when $\left\langle {/s} \right\rangle $ is selected as the next word.

We repeat the decoding process $1,000$ times to sample  $1,000$ outputs for each source sentence.
We additionally add the 1-best output of standard beam search, making the size of the list used for reranking to be $1,001$.

\section{Experiments}
\subsection{Settings}

        \begin{table}[t]\small
        
            \begin{tabular}{l|ll|ll}
              \hline   && &  SOURCE &  TARGET  \\
              \hline   \multirow{5}{0.3in}{en-de}&TRAIN&\#Sents& \multicolumn{2}{c}{1.90M}\\
              &&\#Words& 52.2M&49.7M \\
              &&\#Vocab& 113K&376K \\
              \cline{2-5}
               &DEV&\#Sents&\multicolumn{2}{c}{3,003} \\
                    &&\#Words&67.6K&63.0K\\
               \cline{2-5}
                &TEST&\#Sents&\multicolumn{2}{c}{2,169} \\
                     &&\#Words&46.8K&44.0K\\
              \hline   \multirow{5}{0.3in}{en-fr}&TRAIN&\#Sents& \multicolumn{2}{c}{1.99M}\\
              &&\#Words& 54.4M& 60.4M\\
              &&\#Vocab& 114K&137K \\
              \cline{2-5}
             &DEV&\#Sents&\multicolumn{2}{c}{3,003} \\
             &&\#Words&71.1K&81.1K\\
             \cline{2-5}
               &TEST&\#Sents&\multicolumn{2}{c}{1.5K} \\
                 &&\#Words&27.1K&29.8K\\
            
                \hline   \multirow{5}{0.3in}{en-zh}&TRAIN&\#Sents& \multicolumn{2}{c}{954K}\\
                    &&\#Words& 40.4M& 37.2M\\
                    &&\#Vocab& 504K&288K \\
                    \cline{2-5}
                   &DEV&\#Sents&\multicolumn{2}{c}{2K} \\
                   &&\#Words&77.5K&75.4K\\
                   \cline{2-5}
                     &TEST&\#Sents&\multicolumn{2}{c}{2K} \\
                       &&\#Words&58.1K&55.5K\\
                  
                      \hline   \multirow{5}{0.3in}{en-ja}&TRAIN&\#Sents& \multicolumn{2}{c}{3.14M}\\
                          &&\#Words& 104M& 118M\\
                          &&\#Vocab& 273K& 150K\\
                          \cline{2-5}
                         &DEV&\#Sents&\multicolumn{2}{c}{2K} \\
                         &&\#Words&66.5K&74.6K\\
                         \cline{2-5}
                           &TEST&\#Sents&\multicolumn{2}{c}{2K} \\
                             &&\#Words&70.6K&78.5K\\
                          \hline
            \end{tabular}
        
          \caption{Data sets.}
          \label{data}
        \end{table}

        \begin{table*}[t] \small 
            \center
              \begin{tabular}{l|ll|ll|ll|ll}
              \hline
       &en-zh&&en-ja&&en-de&&en-fr&\\
       &dev&test&dev&test&dev&test&dev&test\\
         
                  \hline
           PBMT&30.73&27.72&35.67&33.46&12.37&13.95&25.96&27.50\\
         
          NMT &34.60&32.71&41.67&39.00&12.52&14.05&23.63&23.99\\
             NMT+lex&36.06&34.80&44.47&41.09&13.36&15.60&24.00&24.91\\
             
          \hline
            NMT+lex+rerank($P_n$) &34.38&33.23&38.92&34.18&12.34&13.59&23.13&23.61\\

             NMT+lex+rerank($S_d$) &36.17&34.09&42.91&40.16&13.08&15.29&24.28&25.71\\

                 NMT+lex+rerank($P_n$+$S_d$)& 37.94&\bf 35.59&45.34&41.75&\bf 14.56&\bf 16.61&\bf 25.96&\bf27.12\\
                      \hline
                      
                          NMT+lex+rerank($P_n$+WP)&37.44&34.93& 45.81& 41.90&13.75&15.46&24.47&25.09\\
                               NMT+lex+rerank($S_d$+WP)&36.44&33.73&43.52&40.49&13.39&15.71&24.74&26.25\\     
                                 NMT+lex+rerank($P_n$+$S_d$+WP)&\bf38.69&\bf35.75&\bf46.92&\bf43.17&\bf14.61&\bf16.65&\bf25.98&\bf27.15\\
                            \hline
                  \end{tabular}
              
                       \caption{Translation results (BLEU). NMT+lex: \cite{arthur-neubig-nakamura:2016:EMNLP2016}; NMT+lex+rerank: we rerank the $n$-best outputs of NMT+lex using different features ($P_n$, $S_d$ and WP). }
                       \label{results}
                
            \center
              \begin{tabular}{l|ll|ll|ll|ll}
              \hline
       &en-zh&&en-ja&&en-de&&en-fr&\\
       &METEOR&chrF&METEOR&chrF&METEOR&chrF&METEOR&chrF\\
         
                  \hline
           PBMT&34.70&37.87&35.22&39.45&26.66&50.02&32.33&56.36\\
         
          NMT &34.51&39.91&35.07&42.02&24.91&44.50&29.58&49.99\\
             NMT+lex&35.56&42.22&36.48&44.34&25.49&45.67&30.10&50.89\\
             
          \hline
            NMT+lex+rerank($P_n$) &34.56&40.80&32.63&38.57&23.57&40.35&29.15&48.64\\

             NMT+lex+rerank($S_d$) &36.02&42.65&36.87&44.85&\bf 26.48&\bf 48.73&\bf 31.56&\bf 54.42\\

                 NMT+lex+rerank($P_n$+$S_d$)&36.40&43.73&37.22&45.69&\bf 26.26&47.27&\bf 31.62&53.99\\
                      \hline
                      
                          NMT+lex+rerank($P_n$+WP)&36.04&42.86&36.90&44.93&25.03&44.05&30.21&50.78\\
                               NMT+lex+rerank($S_d$+WP)&36.34&42.78&37.05&45.03& 26.16&47.82& 31.32&53.75\\     
                                 NMT+lex+rerank($P_n$+$S_d$+WP)&\bf 36.88&\bf 44.09&\bf 37.94&\bf 46.66&\bf 26.20&47.12&\bf 31.61&53.98\\
                            \hline
                  \end{tabular}
              
                       \caption{METEOR and chrF scores  on the test sets for different system outputs in Table~\ref{results}.}
                       \label{otherresults}
                   
                                                         \center
                                                  
                                                           \begin{tabular}{l|ll|ll|ll|ll}
                                                           \hline
                                                        
                                                    &en-zh&&en-ja&&en-de&&en-fr&\\
                                                    &dev&test&dev&test&dev&test&dev&test\\
                                                      
                                                               \hline
                                                   PBMT &1.008&1.018&1.005&0.998&1.077&1.069&0.986&1.004\\
                                                
                                                     NMT &0.953&0.954&0.960&0.961&1.059&1.038&0.985&0.977\\
                                                      NMT+lex  &0.936&0.966&0.955&0.963&1.054&1.019&1.030&0.977\\
                                                                        \hline
                                 NMT+lex+rerank($P_n$) &0.875&0.898&0.814&0.775&0.874&0.854&0.904&0.900\\

                  NMT+lex+rerank($S_d$)   &0.973&0.989&0.985&0.981&1.062&1.060&1.030&1.031\\

                           NMT+lex+rerank($P_n$+$S_d$)  &0.949&0.965&0.945&0.936&1.000&0.992&0.999&0.992\\
                                                                      \hline
                           NMT+lex+rerank($P_n$+WP)&0.996&1.019&0.999&0.983&1.000&0.975&0.998&1.001\\
                                                                    
                             NMT+lex+rerank($S_d$+WP)&1.000&1.024&1.001&1.001&1.011&1.007&0.999&0.989\\
                                                                       
                         NMT+lex+rerank($P_n$+$S_d$+WP)&0.990&1.014&1.000&0.986&1.000&0.989&1.000&0.992\\
                                                                          \hline
                                                               \end{tabular}
                                                           
                     \caption{Ratio of translation length to reference length for different system outputs in Table~\ref{results}.}
                                         \label{ratio}
                                                                  \end{table*}

We evaluated the proposed approach for  English-to-Chinese (en-zh),  English-to-Japanese (en-ja), English-to-German (en-de) and English-to-French (en-fr)    translation tasks.   For the en-zh and en-ja tasks, we used datasets provided for the patent machine         translation task at NTCIR-9 \cite{goto2011overview}.\footnote{Note that NTCIR-9 only contained a Chinese-to-English translation task,   we used English as the source  language in our experiments. In NTCIR-9, the development and test sets were both provided for the zh-en task while only the test set was provided for the en-ja task. We used the sentences from the NTCIR-8 en-ja and ja-en test sets as the development   set in our experiments.}   For the en-de and en-fr tasks, we used  version 7 of the Europarl corpus as training data,   WMT 2014 test sets  as our development sets and WMT 2015 test sets as our test sets.     The detailed statistics for training, development and test sets are given in   Table~\ref{data}.  The word segmentation was done by BaseSeg \cite{zhao2006improved} for  Chinese and  Mecab\footnote{http://sourceforge.net/projects/mecab/files/} for Japanese.      
                                                                                                                            
                        We built attentional NMT systems  with   Lamtram\footnote{https://github.com/neubig/lamtram}.  Word embedding size and hidden layer size are both 512. We used Byte-pair encoding (BPE) \cite{sennrich-haddow-birch:2016:P16-12} and set the vocabulary size to be 50K. We used the Adam  algorithm for optimization.

                        \begin{table*}[t]\small 
                                                 \center
                                                                                             
                                 \begin{tabular}{lp{12cm}}
                          \hline 
                                                                              
                       Source&for \colorbox{light-gray}{hypophysectomized (hypop hy sec to mized)} rats , the drinking water additionally contains 5 \% glucose .\\
                                         \hline 
                                                                              
                   Reference& \begin{CJK}{UTF8}{gbsn}对于(for) \colorbox{light-gray}{去(remove) 垂体(hypophysis)} 大(big) 鼠(rat) ， 饮用水(drinking water) 中(in) 另外(also) 含有(contain) 5 ％ 葡萄糖(glucose) 。\end{CJK}\\
                                        \hline 
                                                                          
                           PBMT& \begin{CJK}{UTF8}{gbsn}用于(for) 大(big) 鼠(rat) \colorbox{light-gray}{垂体(hypophysis) HySecto，(Hy Sec to ，)} 饮用水(drinking water) 另外(also) 含有(contain) 5 ％ 葡萄糖(glucose) 。\end{CJK}\\
                                         \hline 
                                                                           
                     NMT&    \begin{CJK}{UTF8}{gbsn}对于(for) \colorbox{light-gray}{过(pass) 盲肠(cecum)} 的(of) 大(big) 鼠(rat) ， 饮用水(drinking water) 另外(also) 含有(contain) 5 ％ 葡萄糖(glucose) 。\end{CJK}\\
                                      \hline  
                     NMT+lex&  \multirow{3}{5in}{  \begin{CJK}{UTF8}{gbsn}对于(for) \colorbox{light-gray}{低(low) 酪(cheese) 蛋白(protein) 切除(remove)} 的(of) 大(big) 鼠(rat) ， 饮用水(drinking water) 另外(also) 含有(contain) 5 ％ 葡萄糖(glucose) 。\end{CJK}}\\
                                     NMT+lex+$P_n$&\\
                                     NMT+lex+$P_n$+WP&\\
                                  \hline 
                                                                             
                      NMT+lex+$S_d$&   \multirow{3}{5in}{\begin{CJK}{UTF8}{gbsn}对于(for) \colorbox{light-gray}{垂体(hypophysis) 在(is) 切除(remove)} 大(big) 鼠(rat) 中(in) ， 饮用水(drinking water) 另外(also) 含有(contain) 5 ％ 葡萄糖(glucose) 。\end{CJK}}\\
                               NMT+lex+$S_d$+WP&\\
                                     &\\
                      \hline 
                                                                               
                   NMT+lex+$P_n$+$S_d$& \multirow{3}{5in}{ \begin{CJK}{UTF8}{gbsn}对于(for) \colorbox{light-gray}{垂体(hypophysis) 在(is) 切除(remove)} 的(of) 大(big) 鼠(rat) 中(in) ， 饮用水(drinking water) 另外(also) 含有(contain) 5 ％ 葡萄糖(glucose) 。\end{CJK}}\\
                                                                                    
                                  NMT+lex+$P_n$+$S_d$+WP&\\
                                  &\\
                                                                                   
                              \hline

                                    \end{tabular}
                                                                                                        
                              \caption{An example of improving inaccurate rare word translation by using  $S_d$ for reranking.}
                                     \label{rare}
                                                \end{table*}

                   To obtain a phrase-based translation rule table for our forced decoding algorithm, we   used GIZA++ \cite{och2003systematic} and \textit{grow-diag-final-and} heuristic to obtain symmetric word alignments for the training set. Then we extracted the     rule table  using Moses \cite{koehn2007moses}.

               \begin{table*}[t]\small
                    \center
                                                                                                                                                                                                                                                                                
                                                                                                                                                                                                              \begin{tabular}{p{2.8cm}p{12.3cm}}
                                                                                                                                                                                                                \hline
                                                                                                                                                                                                          Source&such changes in reaction conditions include , but are not limited to , \colorbox{light-gray}{an increase in temperature or change in ph} .\\
                                                                                                                                                                                                                 \hline 
                       Reference&\begin{CJK}{UTF8}{gbsn}所(such) 述(said) 反应(reaction) 条件(condition) 的(of) 改变(change) 包括(include) 但(but) 不(not) 限于(limit) \colorbox{light-gray}{温度(temperature) 的(of) 增加(increase) 或(or) pH 值(value) 的(of) 改变(change)} 。\end{CJK}\\
                               \hline 
                       PBMT&\begin{CJK}{UTF8}{gbsn}中(in) 的(of) 这种(such) 变化(change) 的(of) 反应(reaction) 条件(condition) 包括(include) ， 但(but) 不(not) 限于(limit) ， \colorbox{light-gray}{增加(increase) 的(of) 温度(temperature) 或(or) pH 变化(change)} 。\end{CJK}\\
                                \hline 
          NMT&\begin{CJK}{UTF8}{gbsn}这种(such) 反应(reaction) 条件(condition) 的(of) 变化(change) 包括(include) 但(but) 不(not) 限于(limit) \colorbox{light-gray}{pH 或(or) pH 的(of) 变化(change)} 。\end{CJK}\\
                                                                                                                           \hline 
           NMT+lex&\multirow{3}{5in}{\begin{CJK}{UTF8}{gbsn}这种(such) 反应(reaction) 条件(condition) 的(of) 变化(change) 包括(include) ， 但(but) 不(not) 限于(limit) ， \colorbox{light-gray}{pH 的(of) 升高(increase) 或(or) pH 变化(change)} 。\end{CJK}}\\
                                                                                                                                                                                                                                                                                         
                                      NMT+lex+$P_n$&\\
                                                       &\\
                                              \hline 
            NMT+lex+$S_d$&\begin{CJK}{UTF8}{gbsn}这种(such) 反应(reaction) 条件(condition) 的(of) 变化(change) 包括(include) 但(but) 不(not) 限于(limit) ， \colorbox{light-gray}{温度(temperature) 的(of) 升高(increase) 或(or) 改变(change) pH 值(value)} 。\end{CJK}\\
                       \hline 
             NMT+lex+$P_n$+$S_d$&\begin{CJK}{UTF8}{gbsn}这种(such) 反应(reaction) 条件(condition) 的(of) 变化(change) 包括(include) ， 但(but) 不(not) 限于(limit) ， \colorbox{light-gray}{温度(temperature) 的(of) 升高(increase) 或(or) 改变(change) pH 值(value)} 。\end{CJK}\\
                               \hline
                 NMT+lex+$P_n$+WP&\begin{CJK}{UTF8}{gbsn}这种(such) 反应(reaction) 条件(condition) 的(of) 变化(change) 包括(include) ， 但(but) 不(not) 限于(limit) ， \colorbox{light-gray}{pH 的(of) 升高(increase) 或(or) 改变(change) pH 值(value)} 。\end{CJK}\\
                     \hline 
            NMT+lex+$S_d$+WP&\multirow{3}{5in}{\begin{CJK}{UTF8}{gbsn}这种(such) 反应(reaction) 条件(condition) 的(of) 变化(change) 包括(include) ， 但(but) 不(not) 限于(limit) ， \colorbox{light-gray}{温度(temperature) 的(of) 升高(increase) 或(or) 改变(change) pH 值(value)} 。\end{CJK}}\\
                                              NMT+lex+$P_n$+$S_d$+WP&\\
                                    &\\
                              \hline
                                                                                                                                                                                                                                                                                         
                 \end{tabular}
                                                                                                                                                                                                                                                                                                                                                       
                                                                                                                                                                                                                     \caption{An example of improving under-translation and over-translation by using  $S_d$ for reranking.}
                                                                                                                                                                                                                      \label{underover}
                                                                                                                                                                                                                     \end{table*}

                    \subsection{Results and Analysis}
                                                                                                                                                                                                                                                                                                                                                                                                                                            
              Table~\ref{results} shows results of the phrase-based SMT system\footnote{We used the default Moses settings for phrase-based SMT.}, the baseline NMT system, the lexicon integration method \cite{arthur-neubig-nakamura:2016:EMNLP2016} and the proposed reranking method. We tested three features for reranking: the NMT score $P_n$, the forced decoding score $S_d$ and a word penalty (WP) feature, which is the length of the translation.
      The best NMT system and the systems that  have no  significant  difference from the best NMT system at the $p < 0.05$ level using bootstrap resampling  \cite{koehn2004statistical} are shown in bold font.
                                                                                                                                                                                                                                                                                                                                                     
                   As we can see, integrating lexical translation probabilities improved the baseline NMT system and reranking with the three features all together achieved further improvements for all four language pairs. Even on English-to-Chinese and English-to-Japanese tasks, where the NMT system outperformed the phrase-based SMT system by 7-8 BLEU scores, using the forced decoding  score for reranking NMT outputs can still achieve  significant improvements.  With or without the word penalty feature, using both $P_n$ and $S_d$ for reranking gave better results than only using $P_n$ or $S_d$ alone.                                                                                                      
                   We also show METEOR and chrF scores on the test sets in Table~\ref{otherresults}. Our reranking method improved both METEOR and chrF significantly.
                   
                                                \paragraph{The Word Penalty Feature}  
 The word penalty feature generally improved the reranking results, especially when only the NMT score $P_n$ was used for reranking.     As we can see,  using only $P_n$ for reranking decreased the translation quality compared to the standard beam search result of NMT. Because the search spaces of beam search and random sampling are quite different, the best beam search output does not necessarily have the highest NMT score compared to random sampling outputs. Therefore, even the $P_n$ reranking results do have higher NMT scores, but have lower BLEU scores according to Table~\ref{results}. To explain why this happened, we show the ratio of translation length to reference length in Table~\ref{ratio}.  As we can see, the $P_n$ reranking outputs are much shorter. This is because NMT generally prefers shorter translations, since Equation~\ref{p1} multiplies all target word probabilities together. So the word penalty feature can improve the $P_n$ reranking results considerably, by preferring longer sentences.   Because the forced decoding  score $S_d$ as shown in Equation~\ref{p2} does not obviously prefer shorter or longer sentences,  when $S_d$ was used for reranking, the word penalty feature became less helpful. When both $P_n$ and $S_d$  were used for reranking, the word penalty feature only achieved further significant improvement on the English-to-Japanese task.

                                                              \begin{table}[H]\small
                                                                                                    \center
                                                                                                                                                                                              
                                                                                \begin{tabular}{p{6.2cm}r}
                                                                                                                       \hline 
                                                                                                        \bf T$_1$ (NMT+lex):&\\
                                                                                                   {\begin{CJK}{UTF8}{gbsn}for $\to$对于(for)\end{CJK}} &{-3.04}\\
                                                                                              \colorbox{light-gray}{  $r_a$:     \begin{CJK}{UTF8}{gbsn}hy $\to$低(low)\end{CJK}} &\colorbox{light-gray}{-12.19}\\
                                                                             \colorbox{light-gray}{   $r_b$:  \begin{CJK}{UTF8}{gbsn}\texttt{null}$\to$酪(cheese)\end{CJK}} & \colorbox{light-gray}{-21.99}\\
                                                                                       \colorbox{light-gray}{$r_c$: \begin{CJK}{UTF8}{gbsn}\texttt{null}$\to$蛋白(protein)\end{CJK}} & \colorbox{light-gray}{-13.83}\\
                                                                                              {\begin{CJK}{UTF8}{gbsn}to mized $\to$切除(remove)\end{CJK}} &{-6.22}\\
                                                                                                 {\begin{CJK}{UTF8}{gbsn}\texttt{null}$\to$的(of)\end{CJK}} & {-1.53}\\
                                                                                       {\begin{CJK}{UTF8}{gbsn}rats $\to$大(big) 鼠(rat)\end{CJK}} &{-1.52}\\
                                                                                         \colorbox{light-gray}{\begin{CJK}{UTF8}{gbsn}, the drinking water $\to$， 饮用水(drinking water)\end{CJK}} &\colorbox{light-gray}{-1.38}\\
                                                                                               {\begin{CJK}{UTF8}{gbsn}additionally contains $\to$另外(also) 含有(contain)\end{CJK}} &{-3.68}\\
                                                                                                    {\begin{CJK}{UTF8}{gbsn}5 \% $\to$5 ％\end{CJK}} &{-0.51}\\
                                                                                              {\begin{CJK}{UTF8}{gbsn}glucose . $\to$葡萄糖(glucose) 。\end{CJK}} &{-0.60}\\
                                                                                                  \colorbox{light-gray}{ $r_d$: hypop$\to$\texttt{null}} & \colorbox{light-gray}{-25.33}\\
                                                                                                 {sec$\to$\texttt{null}} & {-20.66}\\
                                                                                                                                                                                              
                                                                                       \hline
                                                                                            \bf T$_2$ (NMT+lex+$P_n$+$S_d$):&\\
                                                                                {\begin{CJK}{UTF8}{gbsn}for $\to$对于(for)\end{CJK}} &{-3.04}\\
                                                                               \colorbox{light-gray}{\begin{CJK}{UTF8}{gbsn}hypop hy $\to$垂体(hypophysis)\end{CJK}} &\colorbox{light-gray}{-5.09}\\
                                                                                   \colorbox{light-gray}{\begin{CJK}{UTF8}{gbsn}the $\to$在(is)\end{CJK}} &\colorbox{light-gray}{-5.32}\\
                                                                                    {\begin{CJK}{UTF8}{gbsn}to mized $\to$切除(remove)\end{CJK}} &{-6.22}\\
                                                                                      {\begin{CJK}{UTF8}{gbsn}\texttt{null}$\to$的(of)\end{CJK}} & {-1.53}\\
                                                                                     {\begin{CJK}{UTF8}{gbsn}rats $\to$大(big) 鼠(rat)\end{CJK}} &{-1.52}\\
                                                                                \colorbox{light-gray}{\begin{CJK}{UTF8}{gbsn}, $\to$中(in) ，\end{CJK}} &\colorbox{light-gray}{-4.11}\\
                                                                                \colorbox{light-gray}{\begin{CJK}{UTF8}{gbsn}drinking water $\to$饮用水(drinking water)\end{CJK}} &\colorbox{light-gray}{-1.03}\\
                                                                                        {\begin{CJK}{UTF8}{gbsn}additionally contains $\to$另外(also) 含有(contain)\end{CJK}} &{-3.68}\\
                                                                                               {\begin{CJK}{UTF8}{gbsn}5 \% $\to$5 ％\end{CJK}} &{-0.51}\\
                                                                                                   {\begin{CJK}{UTF8}{gbsn}glucose . $\to$葡萄糖(glucose) 。\end{CJK}} &{-0.60}\\
                                                                                                     {sec$\to$\texttt{null}}& {-20.66}\\
                                                                                                                                                                                                                                               
                                                                                                      \hline
                                                                                                               \end{tabular}
                                                                                                                                                                                                                                                        
                                                                                          \caption{Forced decoding paths for T$_1$ and T$_2$: used rules and log scores. The translation rules with shade are used only for T$_1$ or T$_2$.}
                                                                                                \label{rules}
                                                                                                  \end{table}                                                                                                                                                                                                                                                                                                                                 
                                                                                                  
     Table~\ref{rare} gives translation examples of our reranking method from the English-to-Chinese task.     The source English word ``hypophysectomized" is an unknown word which does not occur in the training set.           By employing BPE, this word is split into ``hypop", ``hy", ``sec", ``to" and ``mized".      The correct translation for ``hypophysectomized" is ``\begin{CJK}{UTF8}{gbsn}去(remove) 垂体(hypophysis)\end{CJK}" as shown in the reference sentence. The original attentional NMT translated it into incorrect translation ``\begin{CJK}{UTF8}{gbsn}过(pass) 盲肠(cecum)\end{CJK}". After integrating lexicons, the NMT system translated it     into ``\begin{CJK}{UTF8}{gbsn}低(low) 酪(cheese) 蛋白(protein) 切除(remove)\end{CJK}". The last word ``\begin{CJK}{UTF8}{gbsn}切除(remove)\end{CJK}" is correct, but the rest of the translation is still wrong.  Only by using the forced decoding  score $S_d$ for reranking, we get the  more accurate translation ``\begin{CJK}{UTF8}{gbsn}垂体(hypophysis) 在(is) 切除(remove)\end{CJK}".

                                                                                                  To further demonstrate how the reranking method works, Table~\ref{rules} shows translation rules and their log-scores contained in the forced decoding paths found for T$_1$, the NMT translation without reranking and T$_2$, the NMT translation using both $P_n$ and $S_d$ for reranking. As we can see, the four rules $r_a$, $r_b$, $r_c$ and $r_d$ used for T$_1$ have  low scores. $r_a$ is an unlikely translation. In $r_b$, $r_c$ and $r_d$, ``\begin{CJK}{UTF8}{gbsn}酪(cheese)\end{CJK}", ``\begin{CJK}{UTF8}{gbsn}蛋白(protein)\end{CJK}" and ``hypop" are content words, which are unlikely to be deleted or inserted during translation.                                  
           Table~\ref{rules} also shows that the translation of function words is very flexible. The  score of inserting a function word ``\begin{CJK}{UTF8}{gbsn}的(of)\end{CJK}" is very high. The translation rule ``\begin{CJK}{UTF8}{gbsn}the $\to$在(is)\end{CJK}" used for T$_2$ is incorrect, but its score is relatively high, because function words are often incorrectly aligned in the training set. The reason why function words are more likely to be incorrectly aligned to each other is that they usually have high frequencies and do not have clear correspondences between different languages.

                                                                                                                                                                                                                                                                                                                                                                                                                                                                    In T$_1$, ``hypophysectomized (hypop hy sec to mized)" is incorrectly translated into ``\begin{CJK}{UTF8}{gbsn}低(low) 酪(cheese) 蛋白(protein) 切除(remove)\end{CJK}". However, from Table~\ref{rules}, we can see that the forced decoding algorithm learns it as unlikely translation (hy$\to$\begin{CJK}{UTF8}{gbsn}低(low)\end{CJK}), over-translation (\texttt{null}$\to$\begin{CJK}{UTF8}{gbsn}酪(cheese)\end{CJK}, \texttt{null}$\to$\begin{CJK}{UTF8}{gbsn}蛋白(protein)\end{CJK}) and under-translation (hypop$\to$\texttt{null}, sec$\to$\texttt{null}), because there is no translation rule between ``hypop" ``sec" and ``\begin{CJK}{UTF8}{gbsn}酪(cheese)\end{CJK}" ``\begin{CJK}{UTF8}{gbsn}蛋白(protein)\end{CJK}". Because content words are unlikely to be deleted or inserted during translation, they have low forced decoding  scores.  So using the forced decoding score for reranking NMT outputs can naturally improve over-translation or under-translation as shown in Table~\ref{underover}. As we can see, without using $S_d$ for reranking, NMT under-translated ``temperature" and over-translated ``ph" twice, which will be assigned low scores by forced decoding. By using $S_d$ for reranking,  the correct translation was selected. 
                                       
                     We  did human evaluation on 100 sentences randomly selected from the English-to-Chinese test set to test the effectiveness of our forced decoding method. We compared the outputs of two systems:   
  \begin{itemize}
  \item NMT+lex+rerank($P_n$+WP)
  \item NMT+lex+rerank($P_n$+$S_d$+WP)     
  \end{itemize}
  For each source sentence, we compared the  two system outputs. Table~\ref{human} shows the numbers of sentences that our forced decoding feature helped to reduce completely unrelated translation, over-translation and under-translation. The last line of Table~\ref{human} means that for 73 source sentences, our forced decoding feature neither reduced  nor caused more unrelated/over/under translation. That is our forced decoding feature never caused more unrelated/over/under translation for the sampled 100 sentences, which shows that our method is very robust for improving   unrelated/over/under translation.

       \begin{table}[!h]\small 
                                           \center
                                                                                                                                                      
                                            \begin{tabular}{ll|r}

                                               \hline 
                                                 \multirow{4}{0.3in}{Reduce}& both under- and over- translation&2\\
                                                 & under-translation&11\\
                                                 & over-translation& 10\\
                                                 & unrelated translation&4\\
                                                   \hline
                                            
                                                                         \multicolumn{2}{l|}{No  difference}  &73\\
                                                                         \hline
                                               \end{tabular}
                                                                                                                                                                                                                                                        
                                               \caption{Human evaluation results.}
                                                    \label{human}
                                                         \end{table}
                                                                                                                                                                                                                                                                                                                                                                                              
                                                                                                                                                                                                          \paragraph{Reranking PBMT Outputs with NMT} We also did experiments that use the NMT score as an additional feature to rerank PBMT outputs (unique $1,000$-best list).                     
        The results are shown in Table~\ref{smt-rerank}. We also copy results of baseline PBMT and NMT  from Table~\ref{results} for direct comparison.        As we can see, using NMT to rerank PBMT outputs achieved  improvements over the baseline PBMT system. However, when the baseline NMT system is significantly better than the baseline PBMT system (en-zh, en-ja), even using NMT to rerank PBMT outputs still achieved  lower translation quality compared to the baseline NMT system.

                              \begin{table}[!h]\small 
                                         \center
                                                                                                                                                    
                                          \begin{tabular}{l|lllll} 
                                             \hline
                                                &  &en-zh&en-ja&en-de&en-fr\\
                                                 \hline
                                           PBMT+rerank& &32.77&37.68&\bf 14.23&\bf 28.86\\
                                             PBMT&dev &30.73&35.67&12.37&25.96\\
                                                  NMT& &\bf 34.60&\bf 41.67&12.52&23.63\\
                                                  \hline
                                           PBMT+rerank & &30.04&35.14&\bf 15.89&\bf 29.77\\

                              PBMT  &test&27.72&33.46&13.95&27.50\\

                                    NMT   &&\bf 32.71&\bf 39.00&14.05&23.99\\
                                                                    \hline
                                             \end{tabular}
                                                                                                                                                                                                                                                      
                                             \caption{Results of using NMT for reranking PBMT outputs.}
                                                  \label{smt-rerank}
                                                       \end{table}                        
\section{Related Work}  
    
    \newcite{wuebker-mauser-ney:2010:ACL,wuebker-hwang-quirk:2012:WMT}   applied forced decoding on the training set  to improve the training process of phrase-based SMT and prune the phrase-based rule table. 
    They also used word insertions and deletions for forced decoding, but they used a high penalty for all insertions and deletions. In contrast, our soft forced decoding algorithm for NMT outputs uses a small penalty for function words and a high penalty for content words, because function words are usually translated very flexibly and more likely to be inserted or deleted compared to content words. For
example, the under-translation of a content word can hurt the adequacy of the
translation heavily. But function words may naturally disappear during translation
(e.g. the English word ``the" disappears in Chinese). By assigning a high penalty  to words that should not be deleted or inserted during translation, our soft forced decoding method aims to improve the adequacy of NMT, which is very different from previous forced decoding methods that are used to improve general SMT training \cite{yu-EtAl:2013:EMNLP,xiao2016loss}.

       A  major difference of traditional SMT and NMT is that the alignment model in traditional SMT  provides exact word or phrase level alignments between the source and target sentences while  the attention model in NMT  only computes an alignment probability distribution for each target word over all source words, which is the main reason why NMT is more likely to produce completely unrelated translations, over-translation or under-translation compared to traditional SMT. To relieve NMT of these problems, there are methods that modify the NMT neural network structure \cite{tu-EtAl:2016:P16-1,meng-EtAl:2016:COLING,alkhouli-EtAl:2016:WMT} while we rerank NMT outputs by  exploiting knowledge from traditional SMT.
       
       There are also existing methods that rerank NMT outputs by using target-bidirectional NMT models \cite{liu-EtAl:2016:N16-11,sennrich-haddow-birch:2016:WMT}. Their reranking method aims to overcome the  issue of unbalanced accuracy in NMT
outputs  while our reranking method aims to solve the  inadequacy problem of NMT.
              
\section{Conclusion}
In this paper, we propose to exploit an existing phrase-based  SMT model to compute the phrase-based decoding cost for  NMT outputs
 and then use the phrase-based decoding cost to rerank the $n$-best NMT outputs, so we can combine the advantages of both PBMT and NMT. 
 Because an NMT output may not be in the search space of standard phrase-based SMT, we propose a forced decoding algorithm, which can always  successfully find  a decoding path for any NMT output by deleting source words and inserting target words.
Results  show that using the forced decoding cost to rerank NMT outputs improved translation accuracy on four different language pairs.

\bibliography{ijcnlp2017}
\bibliographystyle{ijcnlp2017}

\end{document}